\documentclass{article}

\PassOptionsToPackage{numbers, compress}{natbib}

\usepackage[final]{nips_2018}

\usepackage[utf8]{inputenc} 
\usepackage[T1]{fontenc}    
\usepackage{hyperref}       
\usepackage{url}            
\usepackage{booktabs}       
\usepackage{amsfonts}       
\usepackage{nicefrac}       
\usepackage{microtype}      
\usepackage{xcolor}
\usepackage{amsmath,amssymb}
\usepackage{graphicx}
\usepackage{wrapfig}
\usepackage{subfigure}

\usepackage{titlesec}
\titlespacing*{\section}{0pt}{1ex plus .5ex minus .7ex}{.5ex plus .2ex minus .3ex}
\titlespacing*{\paragraph}{0pt}{0ex}{1ex}

\title{Back to square one: probabilistic trajectory 
forecasting without bells and whistles}

\author{
  Ehsan~Pajouheshgar\thanks{Work done while at IST Austria.} \\
  Sharif University of Technology\\
  Tehra, Iran \\
  \texttt{e.pajouheshgar@gmail.com} \\
  \And
  Christoph~H.~Lampert\\
  Institute of Science and Technology Austria (IST Austria)\\
  Klosterneuburg, Austria\\
\texttt{chl@ist.ac.at} \\
}

\usepackage{xspace}
\makeatletter
\DeclareRobustCommand\onedot{\futurelet\@let@token\@onedot}
\newcommand{\@onedot}{\ifx\@let@token.\else.\null\fi\xspace}

\newcommand{\eg}{{e.g}\onedot} 
\newcommand{\ie}{{i.e}\onedot}

\makeatother

\newcommand{\R}{\mathbb{R}}

\begin{document}

\maketitle

\begin{abstract}
  We introduce a spatio-temporal convolutional neural network 
  model for trajectory forecasting from visual sources.
  Applied in an auto-regressive way it provides an explicit
  probability distribution over continuations of a given 
  initial trajectory segment.
  We discuss it in relation to (more complicated) existing
  work and report on experiments on two standard datasets
  for trajectory forecasting: \emph{MNISTseq} 
  and \emph{Stanford Drones}, achieving results 
  on-par with or better than previous methods.
\end{abstract}

\section{Introduction}

\begin{wrapfigure}[19]{r}{.45\textwidth}\centering 
\vspace{-3.5\baselineskip}
\includegraphics[width=.4\textwidth]{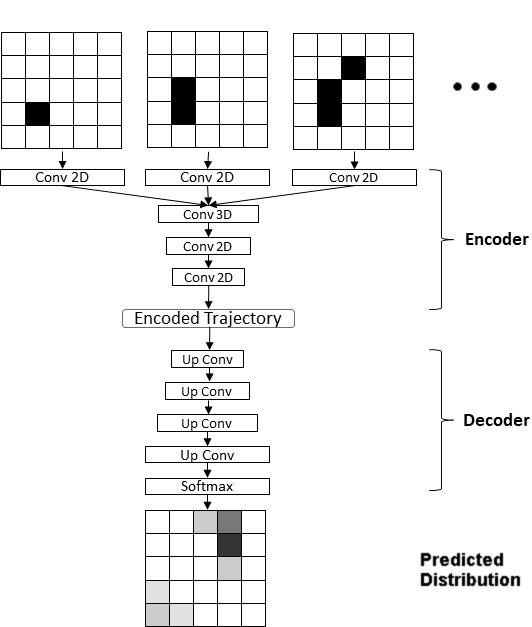}
\caption{Schematic illustration of STCNN (spatio-temporal convolutional neural network)
for probabilistic trajectory forecasting.}
\label{fig:model}
\end{wrapfigure}

A crucial task for an autonomous system, such as a drone or 
a delivery robot, is to react to moving objects in their 
environment, either to avoid them or to interact with them. 
In order to be able to plan such actions, the system needs 
the skill to observe other objects, \eg with cameras, and 
to predict how they are likely to move, \ie forecast their 
trajectories. 
In this work, we study the corresponding spatio-temporal 
learning problem: \emph{trajectory forecasting from visual sources}.
Specifically, we aim at learning not just individual trajectories 
but probability distributions that reflect the intrinsic 
uncertainty of the task. 
 
Existing models for this task typically make use of latent variables to 
achieve the necessary multi-modal output distributions.
Instead, we propose using a spatio-temporal convolutional network 
with a probabilistic output layer. It is simpler and easier 
to train than previous models, yet our experiments show that 
it produces as good or even better forecasts.
We also discuss previously used evaluation measures for trajectory 
prediction, arguing that they are unsatisfactory, because they 
either penalize multi-modal outputs instead of encouraging 
them, or encourage random guessing.
Instead, we propose to use a simple cross-entropy measure that 
is mathematically well-understood and well-defined for all models 
that output discrete probability distributions. 

\section{Probabilistic Trajectory Forecasting}
We formalize the task of probabilistic trajectory forecasting 
as follows:
for any given initial trajectory segment of points, 
$x_{1:s} = (x_1,\dots,x_s)$ with $x_i\in\R^2$ for $i=1,\dots,s$, 
predict a continuation of the trajectory for $T$ more steps, 
$x_{s+1:s+T} = (x_{s+1},\dots,x_{s+T})$, where $T$ is a user-defined 
horizon. 
Reflecting that future trajectories are uncertain, we do not seek
just a deterministic prediction but rather a probability distribution, 
$p(x_{s+1:s+T} | x_{1:s})$, \ie the probability of any 
forecast, conditioned on the observed segment.
Specifically for forecasts from visual sources, we assume that the 
positions correspond to pixel locations in a fixed-size image domain.
Optionally, a reference image, $\textbf{y}$, of the same resolution 
can be given to serve as additional input (conditioning) to the 
prediction task.

As training data, a set of example trajectories is given, 
$\textbf{x}^{(1)},\dots,\textbf{x}^{(n)}$, of 
potentially different lengths, \ie $\textbf{x}^{(i)} = x^{(i)}_{1:T_i} = (x^{(i)}_1,\dots,x^{(i)}_{T_i})$ 
with $x^{(i)}_j\in\R^2$ for any $i=1,\dots,n$ and $j=1,\dots,T_i$. 
When reference images are available, we are given one, $\textbf{y}^{(i)}$, for 
each training trajectory, $\textbf{x}^{(i)}$.
Different trajectories can have the same reference images, 
\eg if there are multiple moving objects in a scene.

\paragraph{A fully convolution spatio-temporal model for probabilitic trajectory forecasting.}

We propose a new model, called STCNN (for 
\emph{spatio-temporal convolutional neural network}),
that combines simplicity, efficiency and expressiveness. 
It has a classical encoder--bottleneck--decoder architecture, 
see Figure~\ref{fig:model}, with only spatio-temporal 
convolutions and upconvolutions (\ie transpose convolutions). 
An initial trajectory segment is encoded visually as a sequence 
of images, which are processed using two-dimensional convolutional layers. 
The information of $s$ time steps is aggregated through a 
three-dimensional convolution layer. This is followed by 
further two-dimensional convolutional layers, the output
of which we consider a latent representation of the 
input segment. 
From the representation, a discrete probability distribution is 
constructed by a sequence of upconvolutions and a 
softmax output layer. Details will be reported in a technical
report, and we will also make our source code publicy available.

For any trajectory segment $\bar x$, the model output  
$g(\bar x)$ is a discrete probability distribution
with as many entries as the image has pixels.
The probability that the next step of the trajectory is at a 
location $x$, we denote as $g(\bar x)[x]$.
From this one-step forecast model we obtain a closed-form expression 
for the probability of any trajectory given an initial segment 
by applying the chain rule of probability:
\begin{align}
p(x_{s+1:s+T} | x_{1:s}) &=\!\!\!\prod_{t=1,\dots,T} p(x_{t+s} | x_{t:t+s-1}) =\!\!\!\prod_{t=1,\dots,T} g(x_{t:t+s-1})[x_{t+s}] 
\end{align}

A major advantage of modeling the probability distribution in 
the output layer, compared to, \eg, a latent variable model, is that 
one can train the model using ordinary maximum likelihood training:
\begin{align}
\max_{\theta}\mathcal{L}(\theta)\quad\text{for}\quad \mathcal{L}(\theta)= \sum_{i=1}^n \sum_{t=1}^{T_i} \log g_{\theta}(x^{(i)}_{t:t+s-1})[x^{(i)}_{t+s}]
\label{eq:loglikelihood-training}
\end{align}
where $\theta$ denotes all parameters of the network model $g$.
Note that at training time, all steps of the trajectories are observed, 
so all terms in~\eqref{eq:loglikelihood-training} can be evaluated 
explicitly. 
In fact, by implementing the model in a fully-convolutional way with 
respect to the time dimension, all terms for one training example 
can be evaluated in parallel with a single network evaluation. 
From the trained model, we can sample trajectory forecasts of arbitrary
length in an auto-regressive way: for each $t=1,\dots,T$, we sample a 
next location $x_{s+t}\sim g(x_{t:t+s-1})$, append it to the existing 
segment, and repeat the procedure at the next location. 

The encoder-decoder architecture also allows an easy integration of 
additional information, in particular about potential reference images. 
Similar to~\cite{walker2016}, we use a standard convolutional network 
to extract a feature representation from the reference image. 
This is concatenated 
with  the latent trajectory representation, such that the upconvolution 
operations can make use of the encoded information both from the trajectory 
and from the reference images.
Again, all terms are observed during training, so both components can 
be trained jointly and efficiently using likelihood maximization.

\paragraph{Comparison to models from the literature.}

Trajectory prediction is a classic computer vision topic,
see~\cite{morris2008} for a survey. 
However, only recent methods are able to produce high-quality 
multi-modal probabilistic outputs. 
Most recent works~\citep{bhattacharyya2018accurate,lee2017desire} 
combine recurrent neural networks, in particular 
LSTMs~\citep{hochreiter1997long}, with conditional variational 
autoencoders (CVAE)~\citep{sohn2015learning}. 
The LSTMs act as deterministic encoder and decoder of trajectories
into and from a latent Euclidean space $\mathcal{Z}$. The CVAE 
models a distribution $q( z' | z )$, where $z,z\in\mathcal{Z}$ 
are latent space encodings of the initial and the forecasted 
trajectory, respectively. 
The CVAEs distribution, $q$, itself is Gaussian with respect to 
$z$, with a neural network providing the mean and covariance 
as functions of $z$. 

From the resulting model one can sample trajectory forecasts: 
one encodes the initial segment, samples latent states according 
to the CVAE distribution and decoding them.
One does not have, however, explicit access to the trajectory
distribution, as this would require an intractable integration 
over a continuous latent space. 
Therefore, one also cannot train the models using actual likelihood 
maximization. Instead, \citep{lee2017desire} minimizes the 
least squares difference between sampled trajectories and 
the ones in the training set plus typical VAE-regularization
terms.  
\citep{bhattacharyya2018accurate} first constructs a sampling-based 
surrogate to the log-likelihood, which, however, is not an 
unbiased estimator, see~\citep{burda2015importance}. 
The actual training objective is another approximation of this with 
an averaging step replaced by a maximization. 

\paragraph{Evaluation measure.}

A priori, it is not clear what is the best evaluation measure for 
probabilistic forecasting, where the goal is to predict a --potentially 
multi-modal-- distribution over possible trajectories, yet a 
test set typically contains only individual realizations.
In this work, we argue in favor of a probabilistically justified 
approach also to the evaluation problem. 
One should use a measure that assigns small values if and only if
exactly the (unknown) target distribution is predicted, and that 
allows a fair comparison between different methods and different 
parametrizations.
A measure that fulfills these properties is the negative cross-entropy,
$D(P;Q)=-\mathbb{E}_{x\sim P} \log Q(x)$, 
between the modeled distribution $Q$ and the target distribution $P$ 
of the test data. 
It can be estimated in an unbiased way from samples 
of the target distribution, as they are available to us in form of 
test data trajectories.
Note that for this expression to be well-defined, in particular 
non-negative and minimal for $P=Q$, the value $Q(x)$ should 
be an actual probability value, not the evaluation of a continuous 
density functional.
This is indeed the case for the discrete model we propose, while 
for CVAE-based models, an additional discretization or binning 
step might be required. 


\paragraph{Evaluation measures in the literature.}

Unfortunately, prior work relied on other 
evaluation measures:
\citep{lee2017desire} reports several different measures, 
based either on the $L^2$ distance between trajectories
at different time-steps, or using a prediction oracle, such 
as measuring the error of only those $10\%$ of predicted 
trajectories that are most similar to the test set trajectories.
\citep{bhattacharyya2018accurate} also reports different values: 
the average $L^2$ distance between full trajectories or 
trajectories at certain time-steps, or the sampling-based 
surrogate to the log-likelihood, which has the form of a 
\emph{softmin} over $L^2$ distances, and therefore improves 
the more sampled trajectories are used to evaluate it.

A shortcoming of these measures is they do not measure 
if the predicted probability reflects 
the true one, not even approximately, but each has a 
bias that allows constructing simple baselines that 
beat the complex learning-based models.
For example, any $L^2$-based distance discourages multi-modal 
output distributions: in the case that multiple future paths 
are possible, the $L^2$ distance is smaller when always 
predicting the mean location instead of different samples 
hitting different modes of the distribution.
The oracle measure does encourage diversity, but it does 
so by simply ignoring all predictions that are not close
to the ground truth. This allows achieving very good scores
by a shot gun type of prediction, where trajectories are 
simply spread out over different directions in the image,
see our discussion below.

\section{Experiments}
\paragraph{Datasets.}

We report on experiments on two datasets that were also used 
in previous work: the \emph{MNIST sequence dataset (MNISTseq)}~\citep{de2016incremental}
consists of the original MNIST digits converted to 
pixel-by-pixel sequential trajectories. While usually 
smooth, trajectories can also have long-distance jumps, 
reflecting the non-trivial topology of the digits.
The \emph{Stanford Drone Dataset (SDD)}~\citep{robicquet2016learning}
contains aerial images of street scenes, with trajectories 
corresponding to the movement of different objects in the scene. 
Because objects move at different speeds, locations at adjacent 
time points can have different distances from each other. 
Long-range jumps are not possible, though.
For SDD, the first image of any sequence is used as a 
reference image. For MNISTseq, no reference images are used.

\begin{figure}\centering
\includegraphics[width=\textwidth]{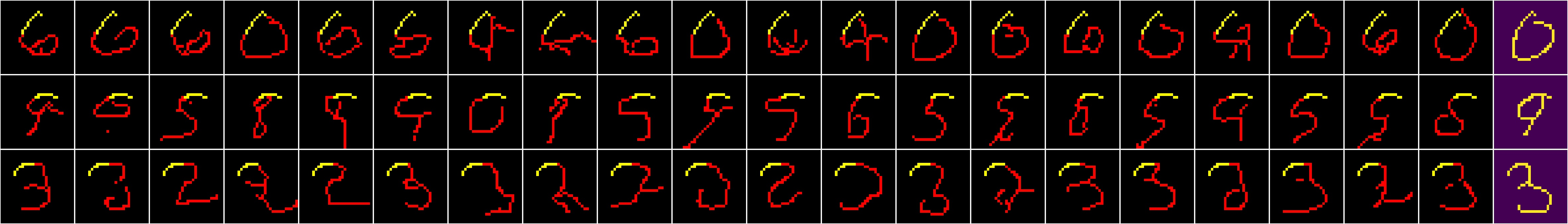}
\includegraphics[width=\textwidth]{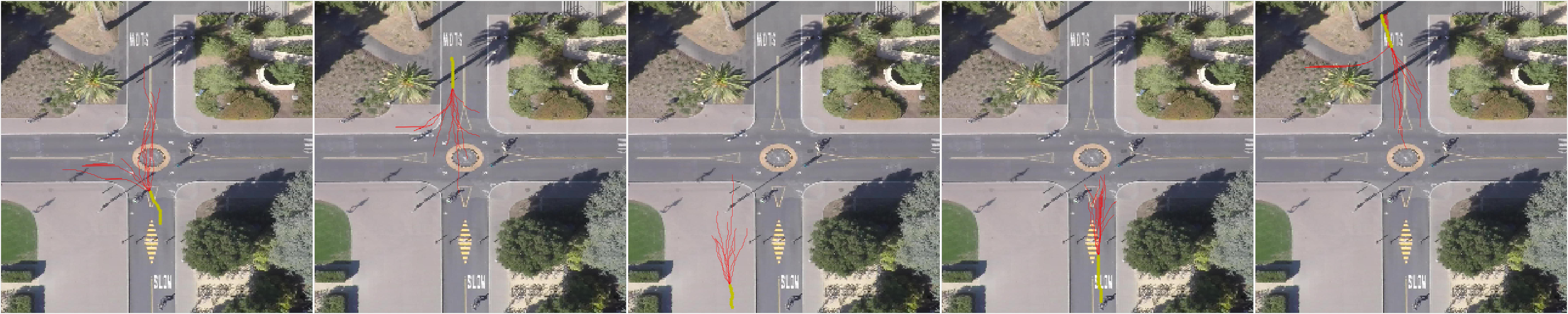}
\caption{Exemplary results on MNISTseq (top rows) and SDD (bottom row); initial segments are depicted in yellow, 
forecasts in red. The learned distributions exhibits multiple modes and also diversity 
within each mode. For MNISTseq, the model is able to also reproduce the occasional jumps 
that occur in the data. For SDD, the model has learned to avoid jumps.}
\label{fig:results}
\end{figure}

\paragraph{Results.}

Qualitative results for our prediction method are depicted in Figure~\ref{fig:results}. 
One can see that the proposed model has learned diverse distributions 
that also reflect specific characteristics of the data, such as the 
presence or absence of jumps. We infer from this that complex 
latent variable models are not required to learn probabilistic and 
multi-modal distribution. A spatio-temporal convolutional structure 
that encodes probability in the output layer is sufficient. 

The quantitative results are promising, too. 
On MNIST, the proposed model achieves a cross-entropy
score of $27.7$, far better than the score of $73.4$ 
achieved by the commonly used baseline of learning a 
Gaussian distribution with mean provided by a LSTM 
that is trained on the sequence coordinates, and with  
Laplacian smoothing. 
Even though~\citep{bhattacharyya2018accurate} also reported result 
on MNISTseq, the values are, unfortunately, not comparable, because 
that model was trained and evaluated on sequences that were padded 
with constant entries to a fixed length. 
Quantitative results for SDD are provided in Table~\ref{tab:resultsSDD} (left). 
Again, STCNN achieves better results than an LSTM baseline. 

\setlength{\tabcolsep}{0.15cm}
\begin{table}\footnotesize
\begin{tabular}{|c|c|c|c|c|}\hline
split & \#samples & STCNN-T & STCNN-T+I & LSTM \\\hline
0 & 3184 & 25.7 & 24.4 & 25.8 \\ 
1	& 1637 &	24.1 & 21.3 & 24.0\\
2	& 2509 &	24.6 & 21.7 & 25.7\\
3	& 4532 &	23.5 & 20.7 & 25.9\\
4 &	3835 &	20.8 & 19.3 & 23.4\\\hline
avg &	15697	& 23.5 $\pm$ 0.8& 21.3 $\pm$ 0.8 & 25.0 $\pm$ 0.5\\\hline
\end{tabular}
\quad
\begin{tabular}{|l|c|c|c|c|}\hline
model &	 \multicolumn{4}{c|}{time point [seconds]}\\
 &	1$s$ & 2$s$ & 3$s$ & 4$s$\\\hline
DESIRE~\citep{lee2017desire} & 1.3	&2.4	&3.5&	5.3\\ 
LSTM-BMS~\citep{bhattacharyya2018accurate} & 0.8	&1.8	&3.1	&4.6\\ 
STCNN (proposed) & 1.2 &	2.1	& 3.3	& 4.6\\ 
"shot gun" baseline & 0.7&	1.7& 3.0& 4.5\\\hline
\end{tabular}
\caption{Quantitative results. Left: negative cross-entropy (mean 
and standard error over five-fold cross-validation) of the
proposed model on SDD with only trajectory information (STCNN-T) or 
with additional reference images (STCNN-T+I). LSTM is a baseline 
architecture, see the main text. Right: top 10\%-oracle 
error at different time points of the forecast, see main text for details.}\label{tab:resultsSDD}
\end{table}

For reference, we also report some measures as they have been used
previously in the literature. Measured by average $L^2$-error, 
our proposed models achieve a value $43.0$, while the (uni-modal) 
LSTM baseline achieves a better value of $35.6$. This is consistent
with the concern that the $L^2$-error penalizes multi-modality 
rather than encouraging it.
For SDD, Table~\ref{tab:resultsSDD} (right) shows the top-10\% oracle error at 
different time points in the forecasted trajectory, 
as in~\citep{bhattacharyya2018accurate,lee2017desire}.
Even though our method yields results comparable to the literature, 
we argue that the oracle measure should not used to judge model quality, 
because it encourages random guessing.
To show this, we performed a simple sanity check: for any initial segment 
we estimate the average speed of the object and extrapolate from the 
last position and orientation in 10 different ways using either slightly 
different orientations $(0^\circ,\pm 8^\circ,\pm 15^\circ)$ or 
different speeds (no movement, or exponentially weighted average 
with coefficients $0, 0.3, 0.7, 1.0$). 
Table~\ref{tab:resultsSDD} (right) shows that this "shot gun" baseline 
achieves oracle scores at least as good as all other methods, despite 
not even making use of the reference image. 

\section{Conclusion}

We proposed a simple probabilistic model for probabilistic 
trajectory forecasting from image data, called STCNN.
It consists of a spatio-temporal convolutional neural networks that 
is applied in an auto-regression way. Probabilistic behavior emerges 
by parametrizing the output layer to represent a (discrete) probability 
distribution. STCNN is easy and efficient to train using a maximum 
likelihood objective and yields promising results. 
Furthermore, we identified shortcomings in the evaluation protocol 
of earlier work, which either discourages or overencourages 
diversity and can be fooled by simple baselines. 
To encourage reproducibility, we will also release our code.

In future work, we plan to reimplement previous models and 
to reevaluate them with the proposed cross-entropy measure to allow fair
comparisons. 
We also plan to study the integration of more contextual 
knowledge, such as trajectories of multiple simultaneously moving 
objects. 

\small
\medskip\noindent\textbf{Acknowledgements.} 
This work was in parts funded by the European Research
Council under the European Union’s Seventh Framework 
Programme (FP7/2007-2013)/ERC grant agreement no 308036.

\bibliographystyle{authordate1}

\bibliography{nips_2018}

\end{document}